\documentclass[10pt,leqno]{amsart}
\usepackage{graphicx}
\usepackage{indentfirst,csquotes}

\usepackage{amssymb,amsthm,amsmath}
\usepackage{xcolor,paralist,hyperref,titlesec,fancyhdr,etoolbox}

\usepackage{graphicx}
\usepackage{amsmath}
\usepackage{amssymb}
\usepackage{booktabs}
\usepackage{dirtytalk}

\usepackage{subcaption}

\usepackage[capitalize]{cleveref}
\crefname{section}{Sec.}{Secs.}
\Crefname{section}{Section}{Sections}
\Crefname{table}{Table}{Tables}
\crefname{table}{Tab.}{Tabs.}

\usepackage[normalem]{ulem}

\def\onedot{\ifx\@let@token.\else.\null\fi\xspace}
\def\eg{\emph{e.g}\onedot} 
\def\eg{\emph{e.g}\onedot} 
\def\ie{\emph{i.e}\onedot} 
 
\def\etc{\emph{etc}\onedot} 
 
\def\etal{\emph{et al}\onedot}
\makeatother

\def\Vec#1{{\boldsymbol{#1}}}
\def\Mat#1{{\boldsymbol{#1}}}

\usepackage{mathtools}
\usepackage[normalem]{ulem}
\usepackage{balance}
\usepackage{amstext}
\usepackage{dsfont}
\balance
\usepackage[linesnumbered,ruled,vlined,onelanguage]{algorithm2e}
\SetKwInput{KwInput}{Input}                
\SetKwInput{KwOutput}{Output} 
\usepackage{authblk} 
\usepackage{xcolor}
\usepackage{babel}

\titleformat{\section}[display]{\normalfont\huge\bfseries\centering}{\centering\chaptertitlename\thechapter}{10pt}{\Large}
\titlespacing*{\section}{0pt}{0ex}{0ex}

\hypersetup{ colorlinks=true, linkcolor=black, filecolor=black, urlcolor=black }

\usepackage{lipsum}
\usepackage{authblk}

\begin{document}
\title{Learning from Noisy Labels with Contrastive Co-Transformer} 

\date{\today}
\email{yan.han@anu.edu.au}
\maketitle

\let\thefootnote\relax
\footnotetext{MSC2020: Primary 00A05, Secondary 00A66.} 

\begin{center}
    {\large \textbf{Yan Han$^{1,3}$, Soumava Kumar Roy$^{1,3}$, Mehrtash Harandi$^{2}$, Lars Petersson$^{3}$}} \\  
    {\small $^1$Australian National University, $^2$Monash University, $^3$CSIRO} \\  
    {\small Emails: \texttt{yan.han@anu.edu.au}, \texttt{soumava.kumar@anu.edu.au}, \\ \texttt{mehrtash.harandi@monash.edu}, \texttt{lars.petersson@data61.csiro.au}}
\end{center}



\begin{abstract}
Deep learning with noisy labels is an interesting challenge in weakly supervised learning. Despite their significant learning capacity, CNNs have a tendency to overfit in the presence of samples with noisy labels. Alleviating this issue, the well known Co-Training framework is used as a fundamental basis for our work. In this paper, we introduce a Contrastive Co-Transformer framework, which is simple and fast, yet able to improve the performance by a large margin compared to the state-of-the-art approaches. We argue the robustness of transformers when dealing with label noise. Our Contrastive Co-Transformer approach  is able to utilize all samples in the dataset, irrespective of whether they are clean or noisy. Transformers are trained by a combination of contrastive loss and classification loss. Extensive experimental results on corrupted data from six standard benchmark datasets including Clothing1M, demonstrate that our Contrastive Co-Transformer is superior to existing state-of-the-art methods. 
\end{abstract} 

\textbf{1. Introduction}
\vspace{0.2cm} 

\label{section:intro}

In this paper, we propose a novel method called Contrastive Co-Transformer (CCT) and demonstrate and its ability to learn from image datasets with noisy labels. Specifically, we add a contrastive module to increase the learning capability of the two homogeneous transformers embedded within the standard Co-Training (Co-Tr) \cite{blum1998combining} learning framework. With such a straightforward augmentation, CCT outperforms the standard baseline algorithms whilst requiring fewer parameters and less computation. We also notice that training of CCT is much more robust compared to standard CNNs as the latter can easily overfit to the noise present in the data.
 
The remarkable successes of Deep Neural Networks can be attributed to manually collecting and annotating large scale datasets (\eg ImageNet~\cite{deng2009imagenet}, Clothing 1M~\cite{liu2016deepfashion} \etc). Due to limited knowledge, understanding of the underlying task and random mistakes, crowd-source workers performing manual annotation of such large scale datasets cannot be expected to annotate with $100$\% accuracy. 
As a result, such corruption of the data leads to incorrect predictions by the deployed model. This highlights the necessity to deal with label noise in an ordered way.

Transformers have recently started to receive significant attention in the field of machine learning and computer vision research.  The seminal work in the field of transformers was proposed in \cite{vaswani2017attention}. Even though the transformers were fundamentally designed to learn long range dependencies in natural language processing \cite{keskar2019ctrl, conneau2019cross, lample2019cross}, they have recently been widely applied to learning discriminative features from images~\cite{liu2021swin, wang2021pyramid, chen2021pre}. The vast amount of research on vision transformer and self-attention based models show the importance of an efficient architectural design that can learn discriminative, distinct, yet complementary visual representations\cite{li2021localvit, srinivas2020curl, yuan2021tokens, wang2021pyramid, zhang2021multi}. As a result of introducing great flexibility in modeling long-range dependencies in vision tasks while introducing less inductive bias, vision transformers have already achieved performances competitive with their CNN counterparts \cite{he2016deep, tan1905rethinking}. They can naturally process multi-modal input data including images, video, text, speech and point clouds. Previous studies \cite{naseer2021intriguing} have shown that transformers are highly robust to severe occlusions. However, there is, to the best of our knowledge, no study discussing the robustness of transformers with respect to label noise. 

Our work studies the robustness to label noise of transformers via experiments on different datasets and settings. We also propose a new Co-Tr method which introduces a contrastive loss module which improves the robustness of the transformer against label noise. 


We begin by finding the importance of introducing transformers, over a standard CNN, as a backbone of Co-Teaching \cite{han2018co} with the help of a toy experiment. Co-Teaching \cite{han2018co} is a well-known method which uses two concurrent CNNs that share information of possibly clean samples during training.  In our toy experiment, we utilise a 3-layer MLP in the case of a CNN based backbone, while using the transformer proposed in~\cite{hassani2021escaping} in the transformer based version. To our surprise, we noticed that transformers are more resilient to noise and are able to reach superior performance in terms of accuracy compared to using an MLP as the backbone of Co-teaching. We set the noise rate of pairflip to $45\%$ (Details of the experiment are attached in the supplementary material). We also observe that CNNs attain a very high accuracy on the test set during the initial epochs, while eventually over-fitting to the noise present in the data. On the other hand, and unlike CNNs, the learning curve of the transformer is slow but steady during the initial epochs, and eventually plateauing as the learning progresses whilst outperforming the CNNs in terms of accuracy on the test set.



Co-Tr is one of the well known algorithms aiming at automatically classifying samples as clean or noisy. Blum~\etal~\cite{blum1998combining} proposed the general Co-Tr learning framework where they successfully addressed the problem of website classification. Recently, Co-Tr has been applied extensively to the noisy label problem\cite{han2018co, wei2020combating, yu2019does}. The main idea behind the Co-Tr algorithm is to simultaneously train two or more sister networks together to learn distinct yet complementary features for the task at hand, thus constraining the need to have two or more different aspects of the data. In our work, we apply transformers to Co-Tr framework and we call it Co-Transformer.

\begin{figure*}[t]
\vspace{-7cm}
\begin{center}
    \includegraphics[width=\linewidth]{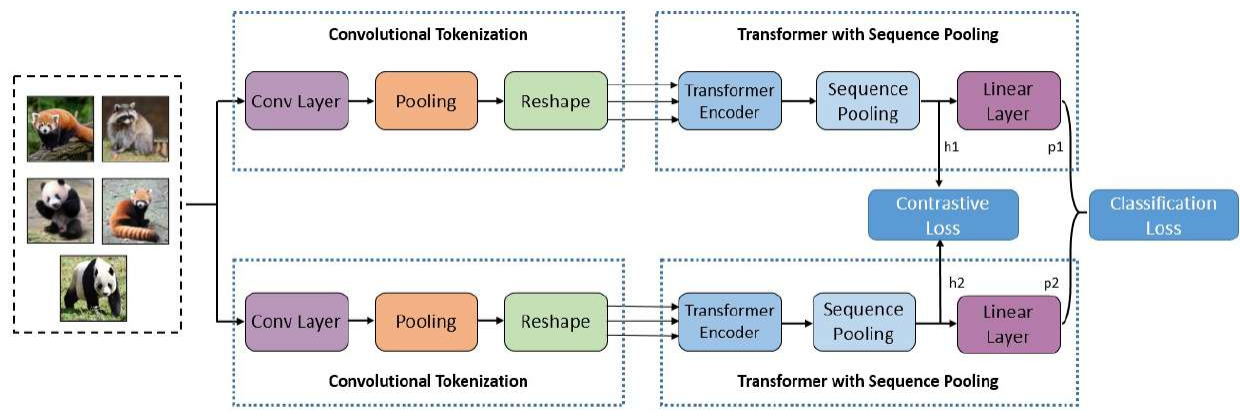}
\end{center}
\vspace{-7cm}
\caption{Schematic diagram of CCT. The noisy dataset is fed into two transformers in parallel. Features ($\Vec{h}_1$ and $\Vec{h}_2$) are extracted before the linear layers. Simultaneously, both $\Vec{h}_1$ and $\Vec{h}_2$ are passed through classifiers for calculating the classification loss ($\Vec{p}_1$ and $\Vec{p}_2$).}
\label{schematic}
\end{figure*}

In order to achieve better performance and robustness against the various degrees of noise present in the data, we also introduce the \emph{contrastive loss} module to the Co-transformer framework.  Contrastive learning on visual representations can be dated back to \cite{hadsell2006dimensionality} and has recently been applied to unsupervised learning \cite{chen2020simple, ye2019unsupervised, doersch2017multi}, which is formulated as the task of finding similar and dissimilar features for a machine learning model. It aims to pull similar examples closer to each other and push dissimilar ones further away which is also needed when learning distinct yet complementary features in a Co-Tr framework in the presence of noisy labels. Generally, for contrastive learning algorithms, data-augmentation techniques are used to create similar examples \cite{chen2020simple,wu2018unsupervised,bachman2019learning,chen2020simple,zhao2020distilling,isola2015learning,hjelm2018learning}. However, for our Co-Transformer we overcome such (in general dataset specific) augmentation techniques by using the outputs of the same input image by Co-transformers as two similar examples. After having extracted the features using two homogeneous transformers, we calculate the contrastive loss for every single example even if the examples have been labeled as clean or noisy by the underlying transformers based on its loss value\footnote{This selection of samples based on their loss value is also famously known as the \say{small-loss} principle.}. As a result, and unlike \cite{han2018co, wei2020combating} which only make use of samples that are labeled as clean, we utilize all the samples within the mini batch in the back propagation of the gradients, resulting in stable and efficient learning. A diagrammatic schematic of our proposed approach is illustrated in Fig \ref{schematic}. 

\begin{figure}[t]
    \centering
    \includegraphics[width=1\textwidth]{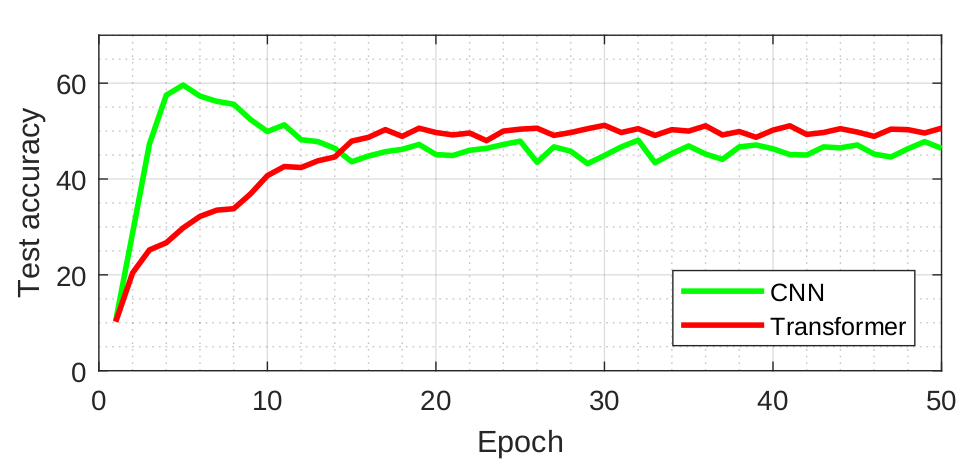}

\vspace{0.1cm}
\caption{Test accuracy of plain CNN and transformer based architectures with $45\%$ of pairflip noise  on CIFAR10.}
\vspace{-12pt}
\label{mlp}
\end{figure}

Our major contributions are:
\begin{itemize}
    \item A novel method to improve the divergence between two homogeneous transformers in the Co-Training framework.
    \item A novel method to extract more information from a noisy mini batch compared to other state-of-the-art methods.
    \item An extensive set of experiments demonstrating the superior performance of our approach and the importance of utilizing all samples in a noisy dataset.
\end{itemize}

\noindent We have evaluated our proposed CCT on six common image classification datasets MNIST~\cite{lecun1998gradient}, CIFAR10~\cite{krizhevsky2009learning},  CIFAR100~\cite{krizhevsky2009learning},  CUBS200-2011~\cite{CUB200_DB},  CARS196~\cite{CARS196_DB} and Clothing1M \cite{xiao2015learning}. Our empirical evaluation clearly demonstrates the effectiveness of CCT over the five state-of-the-art algorithms tested, namely Decoupling~\cite{malach2017decoupling}, F-correction~\cite{patrini2017making}, Mentor-Net~\cite{jiang2018mentornet}, Co-Teaching \cite{han2018co}, JoCor \cite{wei2020combating} and Jo-SRC \cite{yao2021jo} in terms of classification accuracies on the test set of these datasets.










\textbf{2. Related Work}
\vspace{0.2cm} 
\label{gen_inst}

\paragraph{\bf Learning from noisy labels.}
The remarkable success of Deep Neural Networks (DNNs) has made image classification easier in the supervised learning scenario. However, collecting large-scale datasets with precise manual annotations such as ImageNet~\cite{deng2009imagenet} is labour intensive and time consuming. On the other hand, with the growing number of online queries \cite{blum2003noise} and crowd-sourcing \cite{yu2018learning, yan2014learning}, one can easily get vast amounts of information without relying on accurate label annotations. This makes learning with noisy labels an interesting and valuable task. The early methods focus on estimating the noise transition matrix~\cite{patrini2017making, natarajan2013learning, menon2015learning}. However, such an estimation is limited by the number of classes in the dataset. As the number of classes grows, computational complexity of estimation of the noise transition matrix grows quadratically, thereby resulting in slower training time and reduced performance. Instead of such a complex estimation of the noise transition matrix, several other methods~\cite{yu2019does, han2018co} train the network directly with clean and noisy labels. Samples are classified as clean or noisy by the underlying network by inspecting their loss values during training. More precisely, samples with a large loss value are classified as noisy compared to the ones with a lower loss value and vice versa. In recent years, the Co-Training learning framework has become a prominent algorithm when learning from noisy labels~\cite{han2018co, yu2019does, wei2020combating}.

\paragraph{\bf Co-training.}  
Deep Co-Tr was first introduced by \cite{blum1998combining} to classify whether web pages were \say{academic course home pages} or not, using two homogeneous networks. Ever since then, it has been applied in a large variety of different tasks ranging from domain adaptation \cite{chen2019multi}, image classification \cite{qiao2018deep}, data segmentation \cite{chai2011bicos}, tag-based image search \cite{gong2014multi} and many more. Lately, it has been used to tackle the task of learning from samples with noisy labels. \cite{han2018co} successfully applied the Co-Tr framework to select clean samples based on the loss values of the samples during training. More specifically, they follow the principle that examples with a small-loss are more likely to be clean. The first network is trained with the samples that are labelled clean selected by the second network and vice versa; thereby exchanging information between the two using the small-loss clean samples. The same selection strategy has been applied in the work of \cite{yu2019does}, while emphasising how important the disagreement is between the two homogeneous networks used in the learning framework. \cite{wei2020combating} provided a new strategy to learn from noisy labels in order to increase the Co-regularization between two networks. More specifically, they aimed at decreasing the diversity of the entire framework by forcing the output of each network to be similar with each other. Though these methods employ different selection strategies, the underlying notion of annotating small-loss (large-loss) samples as clean (noisy) remains the same; and back propagate the loss from those clean samples only. On the other hand; CCT considers such noisy samples valuable and uses these samples to train the network in an unsupervised manner.

\paragraph{\bf Contrastive Learning.} Contrastive learning as proposed by Hadsell~\etal~\cite{hadsell2006dimensionality} learned representations by contrasting positive pairs against negative pairs. The core idea of contrastive representation learning is to map semantically nearby points (positive pairs) closer together in the embedding space, while pushing apart the points that are dissimilar (negative pairs). Recently, contrastive learning is being used extensively in unsupervised learning~\cite{chen2020simple, kang2019contrastive,srinivas2020curl,arora2019theoretical}. The standard approach uses different augmentation strategies such as horizontal and (or) vertical flips, random crop, colour jitter, random patches \etc to create multiple variations of every individual data point which are then considered positive pairs~\cite{wu2018unsupervised,bachman2019learning,chen2020simple,zhao2020distilling,isola2015learning,hjelm2018learning}. Generating a distinct and an informative positive example for every data point is the fundamental bottleneck in contrastive learning. In our proposed CCT, the features extracted from two encoder networks are considered to be a positive pair. As a result, no additional data augmentation is needed, thereby saving the time and the resources needed to find the right augmentation strategy for every dataset. 

\paragraph{\bf Transformers} Vision Transformers (ViT) were introduced as a way to compete with CNNs on image classification tasks and utilize the benefits of attention within these networks. The motivation was because attention has many desirable properties for a network, specifically its ability to learn long range dependencies. The recent success of ViT \cite{liu2021swin, hassani2021escaping} have provided us with more reliable and excellent performance transformer backbones. \cite{dosovitskiy2020image} argued that Transformers lack some of the inductive biases inherent to CNNs, such as translation equivariance and locality, and therefore do not generalize well when trained on insufficient amounts of data. However, \cite{hassani2021escaping} has shown for the first time that with the right size and tokenization, transformers can perform head-to-head with state-of-the-art methods on small datasets, often with better accuracy and fewer parameters. In this work, we apply transformers from \cite{hassani2021escaping} for small datasets: MNIST, CIFAR10 and CIFAR100. For CU200-2011, CARS196 and Clothing1M we apply transformers from \cite{liu2021swin}.
\textbf{3. Methodology}
\vspace{0.2cm} 

As mentioned above, we aim to utilize all the examples in the noisy datasets and to introduce the contrastive learning algorithm within the Co-Transformer framework. Inspired by the recent success of contrastive learning and Co-training algorithms (as mentioned in the previous section), CCT takes advantages of the small-loss sample selection strategy to boost the diversity when training the two transformers by exchanging information between them. Moreover, samples with labels marked as clean or noisy by the transformers need not to be correct, therefore leading to further erroneous predictions when separating the original noisy samples from the clean ones. The features extracted from the two transformers are different from each other, and thus perfectly fit with the realm of contrastive learning. We therefore add a contrastive loss module in our proposed methodology which considers all the samples within the mini batch, giving a significant performance boost over the current state-of-the-art methods. 

For multi-class classification containing $\mathrm{M}$ classes, a dataset with $\mathrm{N}$ samples is represented as $\mathcal{D} = {\{\Vec{x}_i,\Vec{y}_i\}}^N_{i=1}$, where $\Vec{x}_i$ is the $i^{\textrm{th}}$ instance with its observed label as $\Vec{y}_i \in \{1,...,M\}$. We train networks with mini batches of size $K$.

\textbf{3.1 Transformer}
Based on the conventional Co-Training framework, we formulate the proposed CCT approach with two transformer based encoders denoted by $f(\Vec{x}, \Mat{\Theta}_1)$ and $f(\Vec{x},\Mat{\Theta}_2)$, where $\Mat{\Theta}_1$ and $\Mat{\Theta}_2$ are the parameters of the underlying transformer network respectively. One fundamental advantage of our proposed algorithm is that it is independent of the choice of transformer backbone. Thus, we opt for simplicity and adopt two commonly used transformer backbones from \cite{hassani2021escaping, liu2021swin} (Details of the transformers are shown in Section~\textsection{~\ref{sec:expt}}. We use different configurations of transformers for different datasets, while the configuration of both $f(\Vec{x}, \Mat{\Theta}_1)$ and $f(\Vec{x},\Mat{\Theta}_2)$ remains the same for each separate dataset.) to obtain features $\Vec{h}_i^1 = f(\Vec{x}_i,\Mat{\Theta}_1)$ and $\Vec{h}_i^2 = f(\Vec{x}_i,\Mat{\Theta}_2)$ where $\Vec{h}_i^1, \Vec{h}_i^2 \in \mathbb{R}^d$. We denote the prediction probabilities of $\Vec{x}_i$ as $\Vec{p}_i^1=t(\Vec{h}_i^1,\Mat{\Theta}_1)$ and $\Vec{p}_i^2=t(\Vec{h}_i^2,\Mat{\Theta}_2)$ such that $\Vec{p}_i^1, \Vec{p}_i^2 \in \mathbb{R}^M$ ($t$ is the linear layer.)



\textbf{3.2 Selection Strategy and Classification Loss} Similar to \cite{han2018co,yu2019does, wei2020combating}, we also employ the cross entropy classification loss based selection strategy. Before going into details, we first introduce the relationship between small-loss samples and clean samples. According to \cite{han2018co, wei2020combating}, there is a consensus that samples with a smaller loss are more likely clean, while those samples with larger loss values are more likely to be noisy and corrupted. We obtain the classification loss of the sample $\Vec{x}_i$ for two transformers as shown below:
\begin{equation} 
\label{eq:1-2}
\mathcal{L}_{ce}^1(\Vec{x}_i,\Vec{y}_i)=-\Sigma_{m=1}^M\Vec{y}_i~\textrm{log}(\Vec{p}_1^m(\Vec{x}_i))   ~~~, ~~~\mathcal{L}_{ce}^2(\Vec{x}_i,\Vec{y}_i)=-\Sigma_{m=1}^M\Vec{y}_i~\textrm{log}(\Vec{p}_2^m(\Vec{x}_i))    
\end{equation}

We then select $R$ examples according to $\mathcal{L}_{ce}^1$ and $\mathcal{L}_{ce}^2$ within the mini batch that produces the $R$ lowest cross entropy losses respectively. Thereafter, for each mini batch the classification loss that will be back propagated to update $f(\Vec{x},\Mat{\Theta}_1)$ and $f(\Vec{x},\Mat{\Theta}_2)$ are:
\begin{equation} 
\label{eq:3-4}
\mathrm{L}_{ce}^1=\Sigma_{i=1}^R~\mathcal{L}_{ce}^1(\Vec{x}_i)~~~\forall \Vec{x}_i \in \mathcal{D}_1 ~~~, ~~~
\mathrm{L}_{ce}^2=\Sigma_{i=1}^R~\mathcal{L}_{ce}^2(\Vec{x}_i)~~~\forall \Vec{x}_i \in \mathcal{D}_2
\end{equation}
where $\mathcal{D}_1$ and $\mathcal{D}_2$ represent the set of examples that is used to update the parameters of the network. We follow the work of \cite{han2018co} and two transformers select the $R$ lowest $\mathcal{L}_{ce} $ calculated in Eqn~\ref{eq:1-2} 
for one another. Differently, \cite{wei2020combating} uses the joint classification loss to select examples such that $\mathcal{D}_1 = \mathcal{D}_2$.

\textbf{3.3 Contrastive Loss}
We calculate the conventional contrastive loss as used in~\cite{sohn2016improved,hjelm2018learning,wu2018unsupervised,chen2020simple}. A mini batch of $K$ examples is randomly sampled and we define the contrastive loss on features extracted from two transformers. That is to say, we have $2K$ features. Following the work of \cite{chen2017sampling}, for each sample $\Vec{x}_i$ (or $\Vec{h}_i^1$), we consider the corresponding feature of $\Vec{x}_i$ from the other network (\ie $\Vec{h}_i^2$) as the positive sample and the remaining $(2K-2)$ features as the negative samples. The contrastive loss for the extracted feature $\Vec{h}_i^1$ is defined as below:

\begin{equation}
    \begin{aligned}
    \label{eq:5}
    \mathrm{L}_{con}(\Vec{h}_i^1) &=-\textrm{log}\frac{\textrm{exp}(\textrm{sim}(\Vec{h}_i^1,\Vec{h}_i^2)/\phi)}{\Sigma_{j=1}^{2K}\mathds{1}_{[j\neq i]}\textrm{exp}(\textrm{sim}(\Vec{h}_i^1,\Vec{c}_j)/\phi)} ~~~, 
     \\
    \Vec{c}_{j} &=  \begin{cases}
\Vec{h}_{k}^1 & \text{ if } j\%2==1 
\\ 
\Vec{h}_{k}^2 & \text{ if } j\%2==0 
\end{cases} ~,~  k= \textrm{floor}((j+1)/2)
\end{aligned}
\end{equation} where $\mathds{1}_{[j\neq i]}$ is an indicator function evaluating to 1 when $j\neq i$, $\phi$ denotes a temperature parameter which is set to $0.5$ for all experiments, and \say{$\textrm{sim}$} denotes the standard cosine similarity measure. 
Similarly for $h_i^2$, the contrastive loss is defined as
\begin{equation}
    \label{eqn:7}
    \mathrm{L}_{con}(\Vec{h}_i^2)=-\textrm{log}\frac{\textrm{exp}(\textrm{sim}(\Vec{h}_i^2,\Vec{h}_i^1)/\phi)}{\Sigma_{j=1}^{2K}\mathds{1}_{[j\neq 2i]}\textrm{exp}(\textrm{sim}(\Vec{h}_i^2,\Vec{c}_j)/\phi)}.
\end{equation}
Thus, the overall contrastive loss for a mini batch is given below:
\begin{equation}
    \label{eq:8} 
    \mathrm{L}_{con} =\Sigma_{i=1}^K{[\mathrm{L}_{con}(\Vec{h}_i^1)+\mathrm{L}_{con}(\Vec{h}_i^2)]}
\end{equation}

\textbf{3.4 Comparison and Application}
The final loss of CCT for each transformer is defined as follows:


\begin{equation}
\label{eq:9-10}
 \mathrm{L}_1 = \mathrm{L}_{ce}^1+\lambda~\mathrm{L}_{con} ~~~~~~,~~~~~~ \mathrm{L}_2 = \mathrm{L}_{ce}^2+\lambda~\mathrm{L}_{con},  
\end{equation}
where $\lambda$ is a hyper parameter whose value is fixed to $0.0001$ for every experiment.

Our CCT approach binds the contrastive learning within the noisy learning setting with the result that all examples in a noisy dataset are utilized, such that none of the noisy samples are fully discarded during the training phase. While straightforward and fast, CCT still obtains superior performance over the baseline Co-Tr algorithm. The main differences between CCT and a general Co-Tr framework are as follows: 
\vspace{-7pt}
\begin{enumerate}
    \item CCT makes use of transformers to robustly handle label noise.
    \item CCT treats examples in both a supervised and an unsupervised manner.
    \item When calculating the contrastive loss, all the other $2K-1$ features are considered in calculation of the loss (as shown in Eqn~\ref{eq:5} and~\ref{eqn:7}), thereby utilizing all the samples in the mini batch irrespective of whether they are labeled as clean or corrupted. 
    \item By adding the contrastive loss, the regularization between two transformers increases as it encourages the two extracted features of the same input to be similar; without any introduction of additional learnable parameters in the training phase.
\end{enumerate}


It is to be noted that Wei~\etal~\cite{wei2020combating} also employed \say{contrastive loss} in their proposed method, however there still exist major differences between CCT and~\cite{wei2020combating}. CCT considers all the samples within the mini batch; while Wei~\etal considers one single input from the two base encoder networks and only similarity between positive pairs in the calculation of the contrastive loss. That is to say, for each sample in a mini-batch, CCT's contrastive loss consists of two parts: (a) distance between the two features extracted from the same image (which are minimized); (b) distances between the feature (from one transformer) of the sample and features (from the other transformer) of all other samples in the mini-batch (which are maximized). 
More details are provided in the supplementary material. Moreover, Wei~\etal~\cite{wei2020combating} selects samples based on agreement between the two networks while we consider all the samples within the mini batch in calculation of the contrastive loss irrespective of whether they are labeled as clean or noisy by the underlying transformer based on the \say{small-loss} principle. 

Our proposed CCT is easy to integrate into a conventional Co-Tr framework  while boosting its performance. When dealing with a noisy labelled dataset, the \say{small-loss} principle is widely used \cite{han2018co, wei2020combating,yu2019does}. In different baseline algorithms, various sample selection strategies are used ($\mathcal{D}_i$ in Equation \ref{eq:3-4}). In these small-loss selection strategies, samples with a large cross entropy loss are ignored. However, by using our CCT framework, one does not only keep the principle of selecting clean examples but also utilizing all data when calculating and back propagating the contrastive loss to update the parameters of the underlying networks.




\textbf{4. Experiments}

\vspace{0.2cm} 
\label{sec:expt}


\paragraph{Experimental setup} For results shown in Table~\ref{tab:noise co-teaching}, \ref{tab:noise_cubs} and \ref{tab:clothing}, we assume the noise rate $\tau$ is known. This is a common assumption in recent studies on label noise \cite{han2018co, wei2020combating, yao2021jo}. Please refer to the supplementary material for more details on transformer structure, optimizer and learning rate used for different datasets. Details regarding the noise transition matrices are also provided in the supplementary material.


\begin{table*}[t]
\begin{center}
    \caption{\small Comparison of our proposed CCT against several baseline algorithms. We report the average accuracy (\%) after $5$ runs for CCT. 
    The best and the second best results are shown in \textcolor{red}{\textbf{red}} and \textcolor{blue}{\textbf{blue}} respectively}
    \label{tab:noise co-teaching}
    \vspace{0.2cm}
\scalebox{0.91}{
\begin{tabular}{|c|c|c|c|c|c|c|c|} 
\hline
\multicolumn{1}{|c}{\bf noise} 
&\multicolumn{1}{|c}{\bf Dataset}
&\multicolumn{1}{|c}{\bf F}\cite{patrini2017making}
&\multicolumn{1}{|c}{\bf DC}\cite{malach2017decoupling}
&\multicolumn{1}{|c}{\bf MN}\cite{jiang2018mentornet}
&\multicolumn{1}{|c}{\bf CT}\cite{han2018co}
&\multicolumn{1}{|c|}{{\bf JC}\cite{wei2020combating}}
& \begin{tabular}[c]{@{}c@{}}\textbf{CCT}\\ \textbf{OURS}\\

\end{tabular} 
\\\hline
pairflip-45\% &MNIST &0.24  &58.03 &80.88  &87.63 & \textcolor{blue}{\textbf{93.3}} 
&\textcolor{red}{\bf{94.1}} \\
symmetric-80\% &MNIST &13.35 &28.51 &66.58 &79.73 & \textcolor{blue}{\textbf{84.9}} 
&\textcolor{red}{\bf{85.6} }\\
symmetric-50\% &MNIST &79.61 &81.15 &90.05 &91.32 & \textcolor{blue}{\textbf{95.8}} 
&\textcolor{red}{\bf{96.2}} \\
symmetric-20\% &MNIST &98.82  &95.70 &96.70 &97.2 & \textcolor{blue}{\textbf{97.5}} 
&\textcolor{red}{\bf{97.9}} \\
\hline
pairflip-45\%  &CIFAR10 &6.61  &48.80 &58.14 &72.62 & \textcolor{blue}{\textbf{74.4}} 
&\textcolor{red}{\bf{74.9}} \\
symmetric-80\% &CIFAR10 &15.88 &15.31 &22.15 &26.58  &\textcolor{red}{\textbf{27.8}}  
&\textcolor{blue}{\bf{27.4}} \\
symmetric-50\% &CIFAR10 &59.83 &51.49 &71.10 &74.2 & \textcolor{blue}{\textbf{79.4}} 
&\textcolor{red}{\bf{80.1}} \\
symmetric-20\% &CIFAR10 &84.55  &80.44 &80.76 &82.32 & \textcolor{blue}{\bf{85.7}} 
&\textcolor{red}{\textbf{86.0}} \\
\hline
pairflip-45\% &CIFAR100 &1.60  &26.05 &31.60 &\textcolor{blue}{\textbf{34.8}} & 32.2 
&\textcolor{red}{\bf{34.9}} \\
symmetric-80\% &CIFAR100 &2.1 &3.89 &10.89 &15.2 & \textcolor{blue}{\textbf{15.5}} 
&\textcolor{red}{\bf{15.5}} \\
symmetric-50\% &CIFAR100 &41.04 &25.80 &39.00 &41.4 & \textcolor{blue}{\textbf{43.5}} 
&\textcolor{red}{\bf{44.3}} \\
symmetric-20\% &CIFAR100 &\textcolor{red}{\bf{61.9}}  &44.52 &52.1 &54.2 & 53.0 
&\textcolor{blue}{\textbf{54.2}} \\
\hline
\end{tabular}
}
\end{center}
\vspace{-0.7cm}
\end{table*}

\begin{table}[t]
\centering
    \caption{\small Comparison of our proposed CCT against several baseline algorithms for large fine-grained image recognition datasets for a \textbf{symmetric noise} setup with different values of noise rate $\tau$. We report the average accuracy on the test set (\%) after $5$ runs for CCT. 
    CCT runs with a batch size of $64$).}
    \label{tab:noise_cubs}
    \vspace{0.2cm}
\scalebox{0.9}{

\begin{tabular}{|c|c|c|c|c|c|c|c|c|}
\toprule
 & \multicolumn{4}{c|}{\textbf{CUB200-2011}} & \multicolumn{4}{c|}{\textbf{CARS196}} \\
 \midrule
$\tau$ & \textbf{CE} & \textbf{CT}~\cite{han2018co} & \textbf{JC}~\cite{wei2020combating} & \textbf{\begin{tabular}[c]{@{}c@{}}CCT\\ Ours\end{tabular}} & \textbf{CE} & \textbf{CT}~\cite{han2018co} & \textbf{JC}~\cite{wei2020combating} & \textbf{\begin{tabular}[c]{@{}c@{}}CCT\\ Ours\end{tabular}} \\
\midrule
80\% & 12.3 & 15.2 & 15.9 & \textbf{16.8} & 7.1 & 13.0 & 14.1 & \textbf{15.2} \\
50\% & 43.7 & 55.6 & 56.3 & \textbf{57.2} & 39.7 & 67.7 & 68.3 & \textbf{69.6} \\
20\% & 64.9 & 73.0 & 73.6 & \textbf{74.6} & 72.0 & 86.1 & 86 & \textbf{86.5} \\
\bottomrule
\end{tabular}
}
\end{table}

\textbf{4.1 Comparison with State-of-the-Art}
\paragraph{MNIST.} Rows $1$-$4$ in Table~\ref{tab:noise co-teaching} report the accuracy on the test set of MNIST~\cite{lecun1998gradient}. MNIST is a relatively simple dataset compared to the others considered in our experiments. As observed, in the symmetry case with $\tau=20\%$, which is also the simplest experimental setup, the performance of every algorithm is similar to each other. CCT manages to outperform all the other baseline and state-of-the-art methods, especially when the value of noise rate is high. More specifically, when it comes to more difficult cases, such as symmetric noise with a $\tau$ set to $50\%$ and $80\%$, the performances of F-correction (F)~\cite{patrini2017making}, Decoupling (DC)~\cite{malach2017decoupling} and MentorNet (MN)~\cite{jiang2018mentornet} drop significantly and are not robust to such extreme levels of corruption. However, Co-Teaching (CT)~\cite{han2018co} and JoCoR (JC)\cite{wei2020combating} work well under all these noise rates.  Co-Teaching ~\cite{han2018co} is a clean-sample-selection strategy based method and JoCoR~\cite{wei2020combating} is based on maximizing the agreement between the underlying homogeneous networks. Among all the experimental conditions for MNIST, our CCT achieves the best performance. Especially when the noise setup is $45\%$ pair flip and $80\%$ symmetric, we beat the next best method JoCoR~\cite{wei2020combating} by $0.8\%$ and $0.7\%$ respectively; which is impressive considering those baselines are well over $90\%$ or $80\%$ in terms of test accuracy respectively. 

\paragraph{CIFAR10} Rows $5$-$8$ of Table~\ref{tab:noise co-teaching} show the results of the state-of-the-art methods and our CCT on the noisy CIFAR10~\cite{krizhevsky2009learning} dataset. Similar to MNIST, we outperform the baselines in all the experimental setups except when $\tau$ is set to $20\%$. When $\tau$ is set to $80\%$, CCT still works well but falls behind JoCoR by a negligible value of $0.4\%$. However, for all other noise rates, we consistently achieve the best results. We also observe that F-correction, Decoupling and MentorNet work well when the noise rate is low. With the increase in the noise rate, the performance of F-correction drops dramatically. Decoupling and MentorNet still work well but their performances are not on par with our CCT results.

\paragraph{CIFAR100} The performances of other baseline algorithms and CCT on CIFAR100~\cite{krizhevsky2009learning} dataset are presented in Rows $9$-$12$ of Table~\ref{tab:noise co-teaching}. We note that the performances on CIFAR100 are much lower than MNIST and CIFAR10, as there are $100$ classes in CIFAR100 in comparison to $10$ in MNIST and CIFAR10. We observe that CCT performs better than Co-Teaching and JoCoR (which are relatively robust compared to others) under all evaluated experimental setups. CCT only loses to F-correction when the noise rate is symmetric $20\%$ 

\paragraph{CUBS200-2011} We show the experimental results on CUBS200-2011~\cite{CUB200_DB} in Table~\ref{tab:noise_cubs}. For this dataset, we compare our CCT with three baselines: Cross Entropy classification (CE)\footnote{The network is trained with the cross entropy classification loss using corrupted (or noisy) labels.}, Co-Teaching (CT)~\cite{han2018co} and JoCoR (JC)~\cite{wei2020combating}. Note that, CUBS200-2011 is a more difficult dataset compared to MNIST, CIFAR10 and CIFAR100 as it contains images from 200 fine-grained classes of birds. We observe that our CCT outperforms all the three baseline algorithms considered across several different experimental setups, thus demonstrating that CCT is more robust to various noise conditions against the other methods. 


\paragraph{CARS196}
Table~\ref{tab:noise_cubs} shows results of the performance of CCT on the CARS196~\cite{CARS196_DB} dataset. Similar to the experimental setup of CUBS200-2011, we compare CCT against Cross Entropy (CE), Co-Teaching (CT)~\cite{han2018co} and JoCoR (JC)~\cite{wei2020combating}. CCT outperforms all other baseline methods for several experimental setups. Cross entropy loss fails to perform competitively against the other algorithms when $\tau$ is increased to $50\%$ and $80\%$. In a less challenging setup (\ie $20\%$ symmetric noise), CCT achieves a similar performance as Co-Teaching and JoCoR. However, CCT outperforms both Co-Teaching and JoCoR when $\tau$ is over $50\%$. More specifically, when $\tau$ is $50\%$, CCT beats the next best algorithm JoCoR by $1.3\%$. This is an impressive difference as $50\%$ noise rate on CARS196 is considered to be a very challenging setup. A similar conclusion can be drawn with $\tau=80\%$. 
 
\paragraph{Clothing1M}  
We use the noisy version of the Clothing1M~\cite{xiao2015learning}\footnote{We have used the already noise corrupted version of the Clothing1M dataset in our experiments.} and follow the work of \cite{liu2015classification, yu2018efficient} to infer the noise rate by the clean validation set. The batch size is set to $64$. Further, we report the classification accuracy after training for $51$ epochs for all methods. As shown in Table \ref{tab:clothing}, Our CCT performs better than Co-Teaching~\cite{han2018co} and JoCoR\cite{wei2020combating}; thus demonstrating the effectiveness of CCT over Co-Teaching and JoCoR for a real world corrupted label dataset. More ablation study results can be found in supplementary material.

\begin{table}[t]
    \centering
    \caption{\small Comparison of our proposed CCT against two different baselines on Clothing1M: Co-Teaching~\cite{han2018co} and JoCoR~\cite{wei2020combating}. Accuracy($\%$) is reported as the average of $5$ runs.}
    \label{tab:clothing}
    \vspace{0.3cm}
    \begin{tabular}{|c|c|c|c|}
    \hline
         \multicolumn{1}{|c}{\bf method} &\multicolumn{1}{|c}{\bf Co-Teaching}\cite{han2018co} &\multicolumn{1}{|c}{\bf JoCoR}~\cite{wei2020combating} &\multicolumn{1}{|c|}{\bf CCT(OURS)}\\\hline
         accuracy &68.5 &69.8 &\bf{72.0}\\
         \hline
    \end{tabular}
    
\end{table}

\textbf{5. Conclusion}
\vspace{0.2cm} 

In this paper, we present a straightforward, novel and yet a powerful framework, namely Contrastive Co-Transformer (CCT), which is designed to learn from data with noisy labels. The key motivational idea is to combine contrastive learning with transformers in order to identify clean and noisy labels. Unlike previous methods which only back propagate the gradients based on samples identified as clean, CCT makes use of the entire mini batch of samples using an unsupervised contrastive learning module. Our empirical results on MNIST, CIFAR10, CIFAR100, CUBS200-2011, CARS196 and Clothing1M show that CCT performs well across different complex and very challenging noisy experimental setups. In the future, we aim to integrate and study the effect of several optimal transport based loss functions such as Wasserstein distance, Maximum Mean Discrepancy \etc in the learning framework of CCT.

\bigskip


$\,$

$\,$

\bibliographystyle{plain}
\bibliography{references}



\end{document}